\documentclass[11pt,a4paper]{article}
\usepackage[hyperref]{acl2020}
\usepackage{times}
\usepackage{latexsym}

\usepackage{microtype}
\usepackage{xcolor}
\usepackage{amsmath}
\usepackage{MnSymbol}
\usepackage[tone]{tipa}
\usepackage{CJKutf8}
\usepackage{graphicx}
\newcommand{\ms}[1]{\textcolor{red}{\bf\small }}

\DeclareMathOperator*{\argmax}{argmax}

\aclfinalcopy 
\def\aclpaperid{532} 

\title{Speech Translation and the \textit{End-to-End} Promise:\\ Taking Stock of Where We Are}

\author{Matthias Sperber \\
  Apple \\
  \texttt{sperber@apple.com} \\\And
  Matthias Paulik \\
  Apple \\
  \texttt{mpaulik@apple.com} \\}

\date{}

\begin{document}
\maketitle
\begin{abstract}


Over its three decade history, speech translation has experienced several shifts in its primary research themes; moving from loosely coupled cascades of speech recognition and machine translation, to exploring questions of tight coupling, and finally to end-to-end models that have recently attracted much attention. This paper provides a brief survey of these developments, along with a discussion of the main challenges of traditional approaches which stem from committing to intermediate representations from the speech recognizer, and from training cascaded models separately towards different objectives.

Recent end-to-end modeling techniques promise a principled way of overcoming these issues by allowing joint training of all model components and removing the need for explicit intermediate representations. However, a closer look reveals that many end-to-end models fall short of solving these issues, due to compromises made to address data scarcity. This paper provides a unifying categorization and nomenclature that covers both traditional and recent approaches and that may help researchers by highlighting both trade-offs and open research questions.

\end{abstract}

\section{Introduction}

Speech translation (ST), the task of translating acoustic speech signals into text in a foreign language, is a complex and multi-faceted task that builds upon work in automatic speech recognition (ASR) and machine translation (MT).
ST applications are diverse and include travel assistants \cite{Takezawa1998}, simultaneous lecture translation \cite{Fugen2008}, movie dubbing/subtitling \cite{Saboo2019,Matusov2019a}, language documentation and crisis response \cite{Bansal2017}, and developmental efforts \cite{Black2002}.

Until recently, the only feasible approach has been the cascaded approach that applies an ASR to the speech inputs, and then passes the results on to an MT system. Progress in ST has come from two fronts: general improvements in ASR and MT models, and moving from the loosely-coupled cascade in its most basic form toward a tighter coupling. However, despite considerable efforts toward tight coupling, a large share of the progress has arguably been owed simply to general ASR and MT improvements.\footnote{For instance, \newcite{Pham2019a}'s winning system in the IWSLT 2019 shared ST task \cite{Niehues2019} makes heavy use of recent ASR and MT modeling techniques, but is otherwise a relatively simple cascaded approach.}

Recently, new modeling techniques and in particular end-to-end trainable encoder-decoder models have fueled hope for addressing challenges of ST in a more principled manner. Despite these hopes, the empirical evidence indicates that the success of such efforts has so far been mixed \cite{Weiss2017,Niehues2019}. 

In this paper, we will attempt to uncover potential reasons for this. We start by surveying models proposed throughout the three-decade history of ST. 
By contrasting the extreme points of loosely coupled cascades vs.\ purely end-to-end trained direct models, we identify foundational challenges: 
erroneous early decisions, mismatch between spoken-style ASR outputs and written-style MT inputs, and loss of speech information (e.g.\ prosody) on the one hand, and data scarcity on the other hand. 
We then show that to improve data efficiency, most end-to-end models employ techniques that re-introduce issues generally attributed to cascaded ST.

Furthermore, this paper proposes a categorization of ST research into well-defined terms for the particular challenges, requirements, and techniques that are being addressed or used. This multi-dimensional categorization suggests a modeling space with many intermediate points, rather than a dichotomy of cascaded vs.\ end-to-end models, and reveals a number of trade-offs between different modeling choices.
This implies that additional work to more explicitly analyze the interactions between these trade-offs, along with further model explorations, can help to determine more favorable points in the modeling space, and ultimately the most favorable model for a specific ST application.

\section{Chronological Survey}

This chapter surveys the historical development of ST and introduces key concepts that will be expanded upon later.\footnote{For a good comparison of empirical results, which are not the focus of this paper, we refer to concurrent work \cite{Sulubacak2019}. Moreover, for conciseness we do not cover the sub-topic of simultaneous translation \cite{Fugen2008}.}

\subsection{Loosely Coupled Cascades}
\label{sec:loosely-coupled-cascade}

Early efforts to realize ST \cite{Stentiford1988,Waibel1991}
 introduced what we will refer to as the \textbf{loosely coupled cascade} in which separately built ASR and MT systems are employed and the best hypothesis of the former is used as input to the latter. The possibility of \textbf{speech-to-speech} translation, which extends the cascade by appending a text-to-speech component, was also considered early on \cite{Waibel1991}. 

These early systems were especially susceptible to \textbf{errors propagated} from the ASR, given the widespread use of interlingua-based MT which relied on parsers unable to handle mal-formed inputs \cite{Woszczyna1993,Lavie1996,Liu2003}. Subsequent systems \newcite{Wang1998,Takezawa1998,Black2002,sumita2007nict}, relying on data driven, statistical MT, somewhat alleviated the issue, and also in part opened the path towards tighter integration.

\subsection{Toward Tight Integration}

Researchers soon turned to the question of how to avoid early decisions and the problem of error propagation. While the desirable solution of full integration over transcripts is intractable \cite{Ney1999}, approximations are possible. 
\newcite{Vidal1997,Bangalore2001,Casacuberta2004,perez2007comparison} compute a composition of FST-based ASR and MT models, which approximates the full integration up to search heuristics, but suffers from limited reordering capabilities. A much simpler, though computationally expensive, solution is the \textbf{$n$-best} translation approach which replaces the sum over all possible transcripts by a sum over only the $n$-best ASR outputs \cite{Woszczyna1993,Lavie1996}. 
 Follow-up work suggested \textbf{lattices} and \textbf{confusion nets} \cite{Saleem2004,Zhang2005,Bertoldi2005} as more effective and efficient alternatives to $n$-best lists. Lattices proved flexible enough for integration into various translation models, from word-based translation models to phrase-based ST \newcite{matusov2005phrase,Matusov2008} to neural lattice-to-sequence models \cite{Sperber2017,Sperber2019b,Zhang2019b,Beck2019}.

Another promising idea was to limit the detrimental effects of early decisions, rather than attempting to avoid early decisions. One way of achieving this is to train \textbf{robust translation} models by introducing synthetic ASR errors into the source side of MT corpora \cite{Peitz2012,Tsvetkov2014,Ruiz2015,Sperber2017a,Cheng2018,Cheng2019a}. 
A different route is taken by \newcite{Dixon2011,He2011} who directly optimize ASR outputs towards translation quality.

Beyond early decisions, research moved towards tighter coupling by addressing issues arising from ASR and MT models being trained separately and on different types of corpora. \textbf{Domain adaptation} techniques were used by \newcite{Liu2003,Fugen2008} to adapt models to the spoken language domain. \newcite{Matusov2006,Fugen2008} propose re-segmenting the ASR output and inserting \textbf{punctuation}, so as to provide the translation model with well-formed text inputs. In addition, \textbf{disfluency removal} \cite{Fitzgerald2009} was proposed to avoid translation errors caused by disfluencies that are often found in spoken language.

\newcite{aguero2006prosody,Anumanchipalli2012,Do2017,Kano2018a} propose \textbf{prosody transfer} for speech-to-speech translation by determining source-side prosody and applying transformed prosody characteristics to the aligned target words.

\subsection{Speech Translation Corpora}
\label{sec:e2e-corpora}

It is important to realize that all efforts to this point had used separate ASR and MT corpora for training. This often led to a mismatch between ASR trained on data from the spoken domain, and MT trained on data from the written domain. \textbf{End-to-end ST data} (translated speech utterances) was only available in small quantities for test purposes.

\newcite{Paulik2010} proposes the use of audio recordings of interpreter-mediated communication scenarios, which is not only potentially easier to obtain, but also does not exhibit such domain mismatches. \newcite{Post2013} manually translate an ASR corpus to obtain an end-to-end ST corpus, and show that training both ASR and MT on the same corpus considerably improves results compared to using out-of-domain MT data. Unfortunately, high annotation costs prevent scaling of the latter approach, so follow-up work concentrates on compiling ST corpora from available web sources \cite{Godard2018,Kocabiyikoglu2018,DiGangi2019c,Boito2019,Beilharz2019,Iranzo-Sanchez2019}. Note that despite these efforts, publicly available ST corpora are currently strongly limited in terms of both size and language coverage. For practical purposes, the use of separate ASR and MT corpora is therefore currently unavoidable.

\subsection{End-to-End Models}
\label{sec:survey-e2e}

The availability of end-to-end ST corpora, along with the success of end-to-end models for MT and ASR, led researchers to explore ST models trained in an end-to-end fashion. This was fueled by a hope to solve the issues addressed by prior research in a principled and more effective way.
\newcite{Duong2016,Berard2016,Bansal2018} explore \textbf{direct ST models} that translate speech without using explicitly generated intermediate ASR output.
In contrast, \newcite{Kano2017,Anastasopoulos2018,Wang2020} explore \textbf{end-to-end trainable cascades and triangle models}, i.e. models that do rely on transcripts, but are optimized in part through end-to-end training.
\textbf{Multi-task training} and \textbf{pre-training} were proposed as a way to incorporate additional ASR and MT data and reduce dependency on scarce end-to-end data \cite{Weiss2017,Berard2018,Bansal2019,Stoian2019,Wang2020}.
As these techniques were not able to exploit ASR and MT data as effectively as the loosely coupled cascade, other approaches like \textbf{sub-task training} for end-to-end-trainable cascades \cite{Sperber2019}, \textbf{data augmentation} \cite{Jia2019b,Pino2019a}, knowledge distillation \cite{Liu2019}, and meta-learning \cite{Indurthi2020} were proposed.
\newcite{Salesky2019} propose pre-segmenting speech frames, \cite{Jia2019,Tjandra2019a} explore speech-to-speech translation. \newcite{Sung2019,Gangi2019,DiGangi2019b,Bahar2019b,Inaguma2019,DiGangi2019} transfer ideas from MT and ASR fields to ST. 

\section{Central Challenges}
\label{sec:challenges}

\begin{figure}[tb]
\includegraphics[width=\columnwidth]{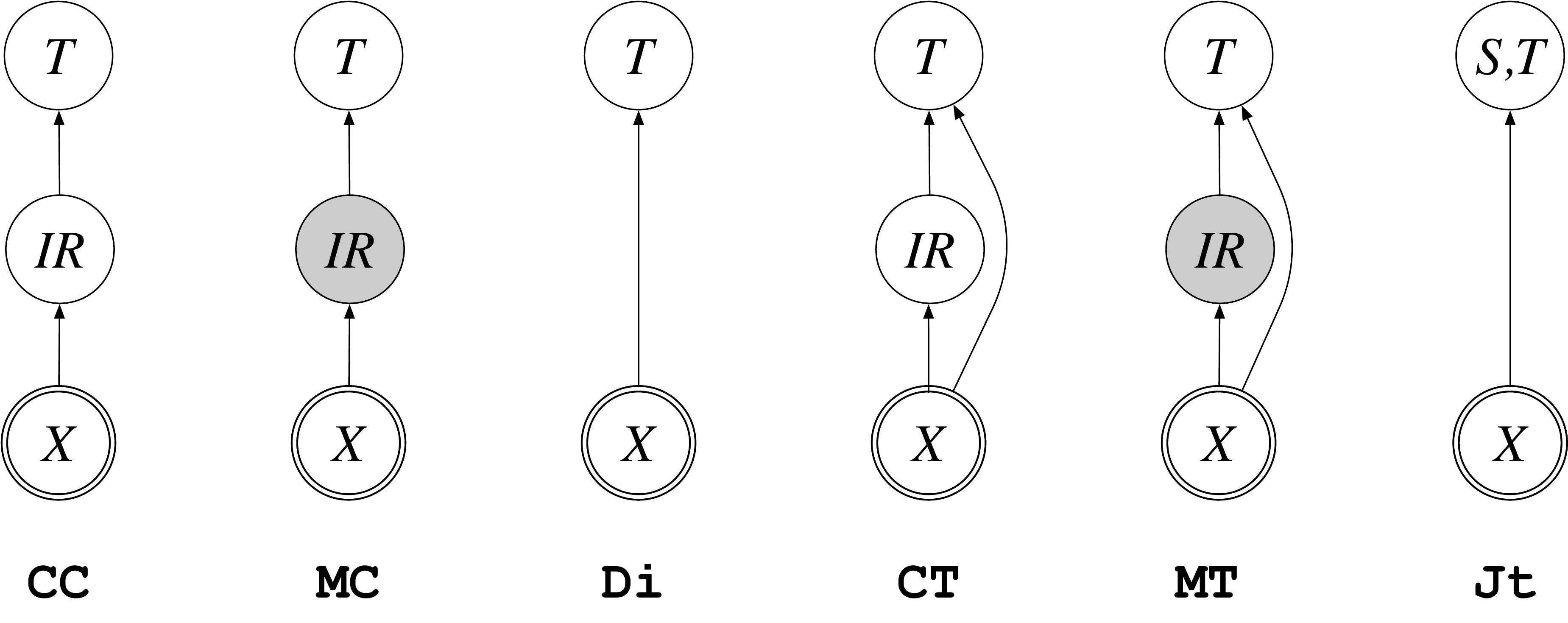}
\caption{Illustration of inference strategies (\S\ref{sec:inference}): Committed/marginalizing cascade (\texttt{CC}/\texttt{MC}), direct (\texttt{Di}), committed/marginalizing triangle (\texttt{CT}/\texttt{MT}), joint (\texttt{Jt}). Double lines differentiate the observed variable (speech input \textit{X}) from random variables (intermediate representations \textit{IR} and translations \textit{T}). Shaded circles marginalize over random variables.}
\label{fig:conditioning}
\end{figure}

Given the abundance of prior work, a clear picture on where we currently stand is needed. For purposes of identifying the key challenges in ST research, this section will contrast the extreme cases of the  \textit{loosely coupled cascade} (\texttt{CC} in Fig.~\ref{fig:conditioning})\footnote{ASR and MT models trained separately on different corpora; intermediate representation is ASR 1-best output.} against the \textit{vanilla direct model} (\texttt{Di} in Fig.~\ref{fig:conditioning}).\footnote{Encoder-decoder model trained on speech utterances paired with translations; no intermediate representations used.} We emphasize that these models are only extreme points in a modeling space with many intermediate points, as we will see in \S\ref{sec:modeling_techniques}. We assume appropriate speech features $X$ as inputs. $T,\hat{T}\in\mathcal{T}$ denote candidate/best translations, respectively, from the MT hypothesis space. $S{\in}\mathcal{H}$ denotes a graphemic transcript from the ASR hypothesis space.

\subsection{Challenges of Loosely Coupled Cascades}
\label{sec:challenges-loosely-coupled}

The loosely coupled cascade justifies its decomposition into MT model $P_\text{MT}\left({T}{\mid}{S}\right)$ and ASR model $P_\text{ASR}\left({S}{\mid}{X}\right)$ as follows:

\begin{align}
\hat{T}&=\argmax_{T\in\mathcal{T}} P\left({T}\mid X\right) \label{eq:cascade1}
\\
&=\argmax_{T\in\mathcal{T}} \sum_{S\in\mathcal{H}}P\left({T}{\mid}{S},{X}\right)P\left({S} {\mid} {X}\right) \label{eq:cascade2}
\\
&\approx\argmax_{T\in\mathcal{T}} \sum_{S\in\mathcal{H}}P_\text{MT}\left({T}{\mid}{S}\right)P_\text{ASR}\left({S}{\mid}{X}\right) \label{eq:cascade3}
\\
&\approx\argmax_{T\in\mathcal{T}} \sum_{S\in\mathcal{H}'}P_\text{MT}\left({T}{\mid}{S}\right)P_\text{ASR}\left({S}{\mid}{X}\right) \label{eq:cascade4}
\end{align}

Note that here the set $\mathcal{H}'$ contains only a single entry, the 1-best ASR output. The approximations in these derivations directly result in the following three foundational challenges:

	\paragraph{Erroneous early decisions:} \textit{Committing to a potentially erroneous $S$ during inference.} This leads to the well-known problem of \textbf{error propagation} \cite{Ruiz2014} and is caused by avoiding the intractable full integration over transcripts (Eq.~\ref{eq:cascade3}) and using only the 1-best ASR output instead (Eq.~\ref{eq:cascade4}). Typical countermeasures include increasing $\mathcal{H}'$ to cover a larger space using lattices or confusion nets, or improving the robustness of MT models.

	\paragraph{Mismatched source-language:} \textit{ASR and MT components model the source-language (transcript) priors }$P_\text{MT}(S)$ \textit{and} $P_\text{ASR}(S)$ \textit{differently.}\footnote{Note that our definition does not entail covariance shift and other forms of domain mismatch \cite{Kouw2018} which, though relevant, are not unique to cascaded ST and are widely covered by general ASR and MT literature \cite{Cuong2018}.} 
	Causes include both modeling assumptions, e.g.\ ASR modeling only unpunctuated transcripts; and mismatched training data, leading to stylistic and topical divergence.
	Typical countermeasures are domain adaptation techniques, disfluency removal, text normalization, and segmentation/punctuation insertion.
	
	\paragraph{Information loss:} \textit{Assumed conditional independence between inputs and outputs, given the transcript: $\big(T\upmodels X\big)\mid{S}$.} This can be seen in Eq.~\ref{eq:cascade3} and results in any information not represented in $S$ to be lost for the translation step. In particular, the MT model is unaware of \textbf{prosody} which structures and disambiguates the utterances, thus playing a role similar to punctuation in written texts; and provides ways to emphasize words or parts of the messages that the speaker think are important. Prosody also conveys information on the speaker’s attitude and emotional state \cite{Jouvet2019}.
	
\begin{table*}[tb]
  \centering
\small
  \begin{tabular}{cc}
    \hline \textbf{English} & \textbf{Japanese} \\ \hline
     			  			& \textit{kochira wa suekko no lucy desu}  \\
    \underline{this} is my \underline{niece} , \underline{lucy} 			  			& \begin{CJK}{UTF8}{min}こちら　は　 姪っ子　の　ルーシー　です　。\end{CJK}  \\
     	&  	\textit{lucy, kono ko ga watashi no suekko desu}  \\
    \underline{this} is my niece , lucy 	&  	\begin{CJK}{UTF8}{min}ルーシー　、　この　子　が　私　の　姪っ子　です　。\end{CJK}  \\
    \hline 
     	 	& \textit{chiizu toka jamu toka, dore ni shimasu ka} \\
    will you have \tone{15}cheese or \tone{15}jam 	 	& \begin{CJK}{UTF8}{min}チーズ　とか　ジャム　とか、　どれ　に　します　か　？\end{CJK} \\
     	 	& \textit{chiizu ka jamu, docchi ni shimasu ka} \\
    will you have \tone{15}cheese or \tone{51}jam 	 	& \begin{CJK}{UTF8}{min}チーズ　か　ジャム、　どっち　に　します　か　？\end{CJK} \\
    \hline
  \end{tabular}
  \caption{Motivating examples for prosody-aware translation from English to Japanese. In the first example, prosody disambiguates whether the speaker is talking about \textit{Lucy} as a third person or directly addressing \textit{Lucy}. In the second example, prosody disambiguates whether \textit{cheese or jam} is an open set or a closed set. In both cases, the surface form of the Japanese translation requires considerable changes depending on the prosody.}
  \label{tab:prosody-examples}
\end{table*}

\subsection{Challenges of the Vanilla Direct Model}
Consider instead the other extreme case: an encoder-decoder model trained to directly produce translations from speech (Eq.~\ref{eq:cascade1}). Because this model avoids the decomposition in Eq.~\ref{eq:cascade2}-\ref{eq:cascade4}, it is not subject to the three issues outlined in \S\ref{sec:challenges-loosely-coupled}.
Unfortunately, this second extreme case is often impractical due to its dependency on scarce end-to-end ST training corpora (\S\ref{sec:e2e-corpora}), rendering this model unable to compete with cascaded models that are trained on abundant ASR and MT training data.

Most recent works therefore depart from this purely end-to-end trained direct model, and incorporate ASR and MT back into training, e.g.\ through weakly supervised training, or by exploring end-to-end trainable cascades or triangle models (\texttt{CT}/\texttt{MT} in Fig.~\ref{fig:conditioning}). This departure raises two questions: (1)~To what extent does the re-introduction of ASR and MT data cause challenges similar to those found in loosely coupled cascades? (2)~Are techniques such as weakly supervised training effective enough to allow competing with the loosely coupled cascade? To address the second question, we propose the notion of data efficiency as a fourth key challenge.

\paragraph{Data efficiency:} \textit{The increase in accuracy achievable through the addition of a certain amount of training data.} To assess data efficiency, data ablations that contrast models over at least two data conditions are required. We argue that empirical evidence along these lines will help considerably in making generalizable claims about the relative performance between two ST models. Generalizable findings across data conditions are critical given that ST models are trained on at least three types of corpora (ASR, MT, and end-to-end corpora), whose availability vastly differs across languages.

\subsection{Data Efficiency vs.\ Modeling Power -- A Trade-Off?}

Consider how the incorporation of MT and ASR data into ST models of any kind may inherently cause the problems as outlined in \S\ref{sec:challenges-loosely-coupled}: Training on MT data may weaken the model's sensitivity to prosody; the effectiveness of training on ASR+MT data may be impacted by mismatched source-language issues; even some types of end-to-end-trainable models make (non-discrete) early decisions that are potentially erroneous.

This suggests a potential trade-off between data efficiency and modeling power. In order to find models that trade off advantages and disadvantages in the most favorable way, it is therefore necessary to thoroughly analyze models across the dimensions of early decisions, mismatched source-language, information loss, and data efficiency.

\paragraph{Analyzing early decisions:}

Problems due to erroneous early decisions are inference-time phenomena in which upstream ASR errors are responsible for errors in the final translation outputs. It follows that the problem disappears for hypothetical utterances for which the ASR can generate error-free intermediate representations. Thus, models that do not suffer from erroneous early decisions will expectedly exhibit an advantage over other models especially for acoustically challenging inputs, and less so for inputs with clean acoustics. This angle can provide us with strategies for isolating errors related to this particular phenomenon. 
Prior work in this spirit has demonstrated that lattice-to-sequence translation is in fact beneficial especially for acoustically challenging inputs \cite{Sperber2017}, and that cascaded models with non-discrete intermediate representations are less sensitive to artificially perturbed intermediate representations than if using discrete transcripts as an intermediate representation \cite{Sperber2019}.

\paragraph{Analyzing mismatched source-language:} End-to-end ST corpora allow for controlled experiments in which one can switch between matched vs.\ mismatched (out-of-domain) MT corpora. \newcite{Post2013} demonstrated that using a matched corpus can strongly improve translation quality for loosely coupled cascades. We are not aware of such analyses in more recent work.

\paragraph{Analyzing information loss:} Prior work \cite{aguero2006prosody,Anumanchipalli2012,Do2017,Kano2018a} has addressed prosody transfer in speech-to-speech translation, but to our knowledge the question of how such information should inform textual translation decisions is still unexplored. Table~\ref{tab:prosody-examples} shows examples that may motivate future work in this direction.

\paragraph{Analyzing data efficiency:} While several prior works aim at addressing this problem, often only a single data condition is tested, limiting the generalizability of findings. We are aware of three recent works that do analyze data efficiency across several data conditions \cite{Jia2019b,Sperber2019,Wang2020}. Findings indicate that both pretraining and data synthesizing outperform multi-task training in terms of data efficiency, and that end-to-end trainable cascades are on par with loosely coupled cascades, while strongly outperforming multi-task training.

\section{Modeling Techniques}
\label{sec:modeling_techniques}

Let us now break apart modeling techniques from prior literature into four overarching categories, with the aim of exposing the ST modeling space between the extreme points of vanilla direct models and loosely coupled cascades.

\subsection{Intermediate Representations}
Almost all models use intermediate representations (IRs) in some form: non-direct models to support both training and inference, and direct models to overcome data limitations. IRs are often speech transcripts, but not necessarily so. A number of factors must be considered for choosing an appropriate IR, such as availability of supervised data, inference accuracy, expected impact of erroneous early decisions, and the feasibility of backpropagation through the IR for end-to-end training. We list several possibilities below:

	\paragraph{Transcripts:} Generally used in the loosely coupled cascade. Being a discrete representation, this option prevents end-to-end training via back-propagation, although future work may experiment with work-arounds such as the straight-through gradient estimator \cite{Bengio2013}. Besides graphemic transcripts, phonetic transcripts are another option \cite{Jiang2011}.
	
	\paragraph{Hidden representations:} \newcite{Kano2017,Anastasopoulos2018,Sperber2019} propose the use of hidden representations that are the by-product of a neural decoder generating an auxiliary IR such as a transcript. Advantages of this representation are differentiability,  prevention of information loss, and weakened impact of erroneous early decisions. A downside is that end-to-end ST data is required for training.
	
	\paragraph{Lattices:} Lattices compactly represent the space over multiple sequences, and therefore weaken the impact of erroneous early decisions. Future work may explore lattices over continuous, hidden representations, and end-to-end training for ST models with lattices as intermediate representation.
	
	\paragraph{Other:} Prior work further suggests pre-segmented speech frames \cite{Salesky2019} or unsupervised speech-unit clusters \cite{Tjandra2019a} as intermediate representation. Further possibilities may be explored in future work.

\subsection{Inference Strategies}
\label{sec:inference}

The conditioning graph (Fig.~\ref{fig:conditioning}) reveals independence assumptions and use of IRs at inference time. Some strategies avoid the problem of early decisions (\texttt{MC}, \texttt{Di}, \texttt{MT}, \texttt{Jt}), while others remove the conditional independence assumption between inputs and outputs (\texttt{Di}, \texttt{CT}, \texttt{MT}, \texttt{Jt}).

	\paragraph{Committed cascade (\texttt{CC}):} \textit{Compute one IR, rely on it to generate outputs (Eq.~\ref{eq:cascade4}).} Includes both the loosely coupled cascade, and recent end-to-end trainable cascaded models such as by \newcite{Kano2017,Sperber2019}.
	\paragraph{Marginalizing cascade (\texttt{MC}):} \textit{Compute outputs by relying on IRs, but marginalize over them instead of committing to one (Eq.~\ref{eq:cascade3}).} As marginalization is intractable, approximations such as $n$-best translation or lattice translation are generally used.
	\paragraph{Direct (\texttt{Di}):} \textit{Compute outputs without relying on IRs (Eq.~\ref{eq:cascade1}).} To address data limitations, techniques such as multi-task training or data augmentation can be used, but may reintroduce certain biases.
	\paragraph{Committed triangle (\texttt{CTr}):} \textit{Commit to an IR, then produce outputs by conditioning on both inputs and intermediate representation.}
	\newcite{Anastasopoulos2018}, who introduce the triangle model, use it in its marginalizing form (see below). Unexplored variations include the use of discrete transcripts as IR, which interestingly could be seen as a strict generalization of the loosely coupled cascade and should therefore never perform worse than it if trained properly.
	\paragraph{Marginalizing triangle (\texttt{MTr}):} \textit{Produce output by conditioning on both input and IR, while marginalizing over the latter (Eq.~\ref{eq:cascade2}).} \newcite{Anastasopoulos2018} marginalize by taking an $n$-best list, with $n$ set to only 4 for computational reasons. This raises the question of whether the more computationally efficient lattices could be employed instead. Similar considerations apply to the end-to-end trainable marginalizing cascade.
	\paragraph{Joint (\texttt{Jt}):} \textit{Changes the problem formulation to }$\hat{S},\hat{T}=\argmax_{S\in\mathcal{H},T\in\mathcal{T}} Pr\left({S,T}\mid X\right)$. This is a useful optimization for many applications which display both transcripts and translations to the user, yet to our knowledge has never been explicitly addressed by researchers.

\subsection{Training Strategies}
\label{sec:inference-strategies}

This group of techniques describes the types of supervision signals applied during \textit{training}.

	\paragraph{Subtask training:} \textit{Training of sub-components by pairing IRs with either the speech inputs or the output translations.} Loosely coupled cascades rely on this training technique while recently proposed cascaded and triangle models often combine subtask training and end-to-end training.
	\paragraph{Auxiliary task training:} \textit{Training by pairing either model inputs or outputs with data from an arbitrary auxiliary task through multi-task training.}\footnote{This definition subsumes pretraining, which is simply using a specific multitask training schedule.} This technique has been used in two ways in literature: (1) To incorporate ASR and MT data into direct models by using auxiliary models that share parts of the parameters with the main model \cite{Weiss2017}. Auxiliary models are introduced for training purposes only, and discarded during inference. This approach has been found inferior at exploiting ASR and MT data when compared to subtask training \cite{Sperber2019}. (2)~To incorporate various types of less closely related training data, such as the use of multitask training to exploit ASR data from an unrelated third language \cite{Bansal2019,Stoian2019}.
	\paragraph{End-to-end:} \textit{Supervision signal that directly pairs speech inputs and output translations.} This technique is appealing because it jointly optimizes all involved parameters and may lead to better optima. The main limitation is lack of appropriate data, which can be addressed by combined training with one of the alternative supervision types, or by training on augmented data, as discussed next.

\subsection{End-to-End Training Data}

	\paragraph{Manual:} \textit{Speech utterances for training are translated (and possibly transcribed) by humans.} This is the most desirable case, but such data is currently scarce. While we have seen growth in data sources in the past two years (\S\ref{sec:e2e-corpora}), collecting more data is an extremely important direction for future work.
	\paragraph{Augmented:} \textit{Data obtained by either augmenting an ASR corpus with automatic translations, or augmenting an MT corpus with synthesized speech.} This has been shown more data efficient than multitask training in the context of adding large MT and ASR corpora \cite{Jia2019b}. \newcite{Pino2019a} find that augmented ASR corpora are more effective than augmented MT corpora. This approach allows training direct models and end-to-end models even when no end-to-end data is available. Knowledge distillation can be seen as an extension \cite{Liu2019}. An important problem that needs analysis is to what extent mismatched source-language and information loss degrade the augmented data.
	\paragraph{Zero-Shot:} \textit{Using no end-to-end data during training.} While augmented data can be used in most situations in which no manual data is available, it suffers from certain biases that may harm the ST model. Similarly to how zero-shot translation enables translating between unseen combinations of source and target languages, 
	it may be worth exploring whether some recent models, such as direct models or cascades with non-discrete IRs, can be trained without resorting to any end-to-end data for the particular language pair of interest.

\section{Applications and Requirements}
\label{sec:requirements}

While we previously described the task of ST simply as the task of generating accurate text translations from speech inputs, the reality is in fact much more complicated. Future work may exploit new modeling techniques to explicitly address the aspects drawn out below.

\subsection{Mode of Delivery}
	\paragraph{Batch mode:} \textit{ A (potentially large) piece of recorded speech is translated as a whole.} Segmentation into utterances may or may not be given. This mode allows access to future context, and imposes no strict computational restrictions. Typical applications include movie subtitling \cite{Matusov2019a} and dubbing \cite{Saboo2019,Federico2020}.
	\paragraph{Consecutive:} \textit{Real-time situation where inputs are provided as complete utterances or other translatable units, and outputs must be produced with low latency.} A typical example is a two-way translation system on a mobile device \cite{Hsiao2006}. This is the only mode of delivery that allows interaction between speaker and translator \cite{Ayan2013}.
	\paragraph{Simultaneous:} \textit{Real-time situation where latency is crucial and outputs are produced incrementally based on incoming audio stream.} Simultaneous translation is faced with an inherent delay vs.\ accuracy trade-off, such as in a typical lecture translation application \cite{Fugen2008}. In addition to computational latency, which is relevant also with consecutive translation, simultaneous translation suffers from inherent modeling latency caused by factors including reordering.

\subsection{Output Medium}

	\paragraph{Text:} This is a standard setting, but is nevertheless worth discussing in more detail for at least two reasons: (1) as is well-known in the subtitling industry, reading speeds can be slower than speaking and listening speeds \cite{Romero-fresco2009}, implying that a recipient may not be able to follow verbatim text translations in case of fast speakers, and that summarization may be warranted. (2) Text display makes repair strategies possible that are quite distinct from spoken outputs: One can alter, highlight, or remove past outputs. One possible way of exploiting this is \newcite{Niehues2018}'s strategy of simultaneous translation through re-translation.
	\paragraph{Speech:} Speech outputs have been used since the early days \cite{Lavie1996}, but whether to apply text-to-speech on top of translated text has often been seen as a question to leave to user interface designers. Here, we argue that ST researchers should examine in what ways speech outputs should differ from text outputs. For example, is disfluency removal \cite{Fitzgerald2009} beneficial for speech outputs, given that human listeners are naturally able to repair disfluencies \cite{lickley1994detecting}? Further examples that need more exploration are prosody transfer \cite{aguero2006prosody} and models that directly translate speech-to-speech \cite{Jia2019}.

\subsection{The Role of Transcripts}

	\paragraph{Mandatory transcripts:} \textit{User interface displays both transcripts and translations to the user.} This scenario has been implemented in many applications \cite{Hsiao2006,Cho2013a}, but has received little attention in the context of end-to-end ST research. It ties together with the \textit{joint} inference model (\S\ref{sec:inference-strategies}). Note that with loosely coupled cascades, there is little need to consider this scenario explicitly because the application can simply display the by-product transcripts to the user. But this is not easily possible with direct models or with models using IRs other than transcripts.

	\paragraph{Auxiliary transcripts:} \textit{Transcriptions are not needed as user-facing model outputs, but may be exploited as IRs during training and possibly inference.} This is the most typical formal framing of the ST task, assuming that transcribed training data is useful mainly for purposes of improving the final translation.

	\paragraph{Transcript-free:} \textit{No transcribed training data exists, so the model cannot rely on supervised transcripts as IR.} The main scenario is endangered language preservation for languages without written script, where it is often easier to collect translated speech than transcribed speech \cite{Duong2016}.

\subsection{Translation Method}
\begin{table}
\small
\centering
\begin{tabular}{cc}
\hline 
ES  & también tengo um eh estoy tomando una clase ..  \\
EN  & i also have um eh i’m taking a marketing class ..  \\
\hline
ES  &  porque qué va, mja ya te acuerda que ..  \\
EN  & because what is, mhm do you recall now that ..  \\
\hline
\end{tabular}
\caption{\label{tab:ex-faithful} Examples for faithful Spanish to English translations, taken from \cite{Salesky2019a}.}
\end{table}

The method of translation is an especially relevant factor in ST, which commonly includes a transfer from the spoken into the written domain. Here, we provide two reference points for the method of translation, while referring to \newcite{Newmark1988} for a more nuanced categorization. 
\paragraph{Faithful:} \textit{Keeps the contextual meaning of the original as precisely as possible within the grammatical constraints of the target language.} With text as output medium, faithful translation may result in poor readability, e.g.\ due to the translation of disfluencies (Table~\ref{tab:ex-faithful}). Arguably the most appropriate output medium for faithful ST would be speech, although user studies are needed to confirm this. Another application are high-stake political meetings in which translations must stay as close to the original sentence as possible. As we move toward more distant language pairs, the practicability of faithful translation of spoken language with disfluencies becomes increasingly questionable.
\paragraph{Communicative:} \textit{Renders the contextual meaning of the original such that both content and style are acceptable and comprehensible by the target audience.} An important example for improving communicativeness is disfluency removal \cite{Fitzgerald2009}. Given that human translators and interpreters adapt their translation method depending on factors that include input and output medium \cite{He2016c}, more research is needed beyond disfluency removal. Communicative translations are especially relevant in casual contexts where convenience and low cognitive effort are mandative. Arguably the closest neighbor of spoken language style in the text realm is social media, it would be interesting to attempt speech-to-text translation with social-media style outputs.

\section{Discussion}

Recent works on end-to-end modeling techniques are motivated by the prospect of overcoming the loosely coupled cascade's inherent issues, yet of the issues outlined in \S\ref{sec:loosely-coupled-cascade}, often only the goal of avoiding early decisions is mentioned motivationally. While early decisions and data efficiency have been recognized as central issues, empirical insights are still limited and further analysis is needed. Mismatched source-language and information loss are often not explicitly analyzed.

We conjecture that the apparent trade-off between data efficiency and modeling power may explain the mixed success in outperforming the loosely coupled cascade. In order to make progress in this regard, the involved issues (early decisions, mismatched source-language, information loss, data efficiency) need to be precisely analyzed (\S\ref{sec:challenges}), and more model variants (\S\ref{sec:modeling_techniques}) should be explored. As a possible starting point one may aim to extend, rather than alter, traditional models, e.g.\ applying end-to-end training as a fine-tuning step, employing a direct model for rescoring, or adding a triangle connection to a loosely coupled cascade. We further suggest that more principled solutions to the different application-specific requirements (\S\ref{sec:requirements}) should be attempted. Perhaps it is possible to get rid of segmentation as a separate step in batch delivery mode, or perhaps text as output medium can be used to visualize repairs more effectively. Several of the application-specific requirements demand user studies and will not be sufficiently solved by relying on automatic metrics only.

\section{Conclusion}
We started this paper with a chronological survey of three decades of ST research, focusing on carving out the key concepts.
We then provided definitions of the central challenges, techniques, and requirements, motivated by the observation that recent work does not sufficiently analyze these challenges. We exposed a significant space of both modeling ideas and application-specific requirements left to be addressed in future research.

Our hope is to encourage meaningful and generalizable comparisons on our quest toward overcoming the long-standing issues found in ST models.



\bibliography{library}
\bibliographystyle{acl_natbib}

\end{document}